\documentclass[review]{elsarticle}
\usepackage{graphicx}
\usepackage{subcaption}
\usepackage{amsmath,amssymb} 
\usepackage{color}
\usepackage{tabu}
\usepackage{multirow} 
\usepackage{adjustbox}
\usepackage{multirow} 
\usepackage{filecontents}
\usepackage{lineno}
\usepackage{booktabs}
\usepackage{siunitx}
\modulolinenumbers[5]

\journal{Journal of \LaTeX\ Templates}
\newcommand{\minisection}[1]{\vspace{0.04in} \noindent {\bf #1}\ \ }

\begin{document}

\begin{frontmatter}

\title{Saliency for Fine-grained Object Recognition in Domains with Scarce Training Data}

\author[mymainaddress,mysecondaryaddress]{Carola Figueroa Flores\corref{mycorrespondingauthor}}
\cortext[mycorrespondingauthor]{Corresponding author}
\ead{cafigueroa@cvc.uab.es}

\author[mymainaddress]{Abel Gonzalez-Garcia}
\author[mymainaddress]{Joost van de Weijer}
\author[mymainaddress]{Bogdan Raducanu}

\address[mymainaddress]{Computer Vision Center\\
Edifici ``O'' - Campus UAB\\
08193 Bellaterra (Barcelona), Spain}

\address[mysecondaryaddress]{Department of Computer Science and  Information Technology\\ Universidad del B\'io  B\'io, Chile}



\begin{abstract}
This paper investigates the role of saliency to improve the classification accuracy of a Convolutional Neural Network (CNN) for the case when scarce training data is available. Our approach consists in adding a saliency branch to an existing CNN architecture which is used to modulate the standard bottom-up visual features from the original image input, acting as an attentional mechanism that guides the feature extraction process. The main aim of the proposed approach is to enable the effective training of a fine-grained recognition model with limited training samples and to improve the performance on the task, thereby alleviating the need to annotate a large dataset.
The vast majority of saliency methods are evaluated on their ability to generate saliency maps, and not on their functionality in a complete vision pipeline. Our proposed pipeline allows to evaluate saliency methods for the high-level task of object recognition. We perform extensive experiments on various fine-grained datasets (Flowers, Birds, Cars, and Dogs) under different conditions and show that saliency can considerably improve the network's performance, especially for the case of scarce training data. Furthermore, our experiments show that saliency methods that obtain improved saliency maps (as measured by traditional saliency benchmarks) also translate to saliency methods that yield improved performance gains when applied in an object recognition pipeline.

\end{abstract}


\end{frontmatter}


\section{Introduction}


Fine-grained object recognition focuses on the classification of subclasses belonging to the same category. Examples of fine-grained datasets include natural categories such as flowers~\cite{nilsback2008automated}, birds~\cite{WelinderEtal2010}, dogs \cite{KhoslaYaoJayadevaprakashFeiFei_FGVC2011} and man-made categories such as cars \cite{krause2013}.  The problem of fine-grained object classification is difficult because the differences between subclasses are often subtle and expert labelers, with knowledge of the discriminating attributes, are needed for the collection of datasets.
Therefore the collection of large datasets is expensive and the development of algorithms that only require few labeled examples is of special interest to the field. 

Computational saliency estimation aims to identify to what extent regions or objects stand out with respect to their surroundings to human observers. Saliency methods can be divided into methods that aim to identify the salient object (or objects) and methods that aim to produce a saliency map that is in according to measurements of human eye-movements on the same image. Itti et al.~\cite{itti1998model} proposed one of the first computational saliency methods based on combining the saliency cues for color, orientation and luminance. Many works followed proposing a large variety of hand-crafted features for saliency~\cite{Ramanathan2010,Borji2014}. Similar as other fields in computer vision, computational saliency estimation has moved in recent years from hand-designed features to end-to-end learned deep features~\cite{Li2016deepsaliency}. 

Saliency detection in human vision plays a role in the efficient extraction of information by placing the attention on those regions in the image that are most informative. However, the vast majority of saliency methods are not evaluated on their efficiency to improve object recognition but instead are evaluated on the task of how accurate their generated saliency masks are. Given that saliency is only an intermediate step of the visual pipeline, evaluating the efficiency of saliency in terms of an improvement of the final task - here we consider fine-grained recognition - could be considered a more valuable evaluation. Therefore, in this paper we aim to evaluate the usefulness of saliency by directly evaluating its improvement on image classification. 

Previous works have found that the incorporation of attention mechanisms in neural networks could be beneficial. 
This theory was subsequently extended to captioning methods where the attention highlights the part of the image that is currently being described by words. Similar to these methods we will incorporate a saliency model, which modulates the normal forward pipeline similarly as an attention model would, but now within the context of fine-grained image classification. Contrarily to these attention methods, we use a saliency network that is pretrained on the task of saliency estimation. Especially, we are interested in demonstrating its effectiveness in the case of scarce training data, a scenario where attending to the relevant information from the image can significantly reduce the danger of overfitting. The main underlying idea is that using saliency as an attention mechanism can help backpropagation to focus on the relevant image information; something which is especially important when only few training examples are available.

In this paper, we investigate to what extent saliency estimation can be exploited to improve the training of an object recognition model when scarce training data is available. 
For that purpose we design an image classification deep neural network that incorporates saliency information as input.
This network processes the saliency map through a dedicated network branch and uses the resulting features to modulate the standard bottom-up visual features from the original image input. 
The main aim of the proposed method is to enable the effective training of a fine-grained recognition model with limited training samples and to improve the performance on the task, thereby alleviating the need to annotate a large dataset. 
We evaluate our method on different datasets and under different settings, achieving considerable performance improvements when leveraging saliency data, especially when training data is scarce.

This paper is organized as follows. In section~\ref{sec:relatedwork} we discuss the related work. 
Section~\ref{sec:method} describes our method in detail, and we perform extensive experiments in section~\ref{sec:experiments}.
Finally, section~\ref{sec:conclusions} presents the conclusions.

\section{Related Work}
\label{sec:relatedwork}

\minisection{Saliency estimation:} 
The seminal work of Itti et al.~\cite{itti1998model} proposed one of the first biologically motivated computational models for saliency estimation. Their saliency map was inferred from multi-scale representations of color, orientation and intensity contrast. Saliency research was propelled further by the availability of large data sets which allowed for direct comparison of methods and enabled the use of data-driven methods based on machine learning algorithms.
The question of whether saliency is important for object recognition has been raised in \cite{vasconcelos2010saliency}. Using a biologically plausible mechanism, the authors demonstrated that indeed saliency (of a top-down nature) has a positive impact on classification. Besides object recognition, saliency has also been used for object tracking.
In \cite{vasconcelos2009saliency}, the authors formulated discriminant tracking as a saliency problem and addressed it using a biologically inspired framework.

Recent methods in saliency are mostly based on deep learning networks. 
Initially, pretrained deep convolutional networks were used directly to extract features for saliency estimation. 
Afterwards, end-to-end networks that learn a mapping from the input image to the saliency map~\cite{Li2016deepsaliency} were introduced. 
But like most previous work on saliency estimation, the main focus of these works is to estimate a saliency map, not how saliency could contribute in a object recognition pipeline. 
In this paper, we aim to investigate if saliency can improve the recognition of objects with deep neural networks. 

\minisection{Attention:} 
The method proposed in this paper is partially based on insights gained from some recent work on attention in neural networks. In~\cite{wang2017residual} the authors propose a method that incorporates attention branches within a feed-forward network for object classification. The attention map, which is repeated for multiple layers in the network, learns to modulate the network features with an attention mechanism. Our saliency branch is similar to the proposed attention mechanism in~\cite{wang2017residual}. 
In our work, however, we use a pretrained saliency branch that is optimized to return a saliency map in accordance with human vision. 
The fact that the network is pretrained is important because that allows it to be used even for object classification problems with very few training examples. 
In this case, the proposed method in~\cite{wang2017residual} would probably fail because it would have to train additional parameters for the attention branch, which would be extremely challenging in the scarce data domain.

Zagoruyko and Komodakis~\cite{Zagoruyko2017} propose a method to train a student network from a teacher network. Their novelty with respect to earlier work is the usage of attention to guide the teaching of the student network. They construct a spatial attention map by considering the activations of an image in a teacher network, and mapping these activations to a single spatial attention map which reflects on what locations the hidden neuron activations were most prominent. They consider that this information is important and can help guide the training process once it is also transfered to the student network. They show that their approach significantly improves the learning of the student network. 
Concretely, they show that guiding the backpropagation of gradients by telling to what spatial coordinates to `attend' can assist in the training process. 
Our paper supports this claim by showing that spatial guidance can help training, although within a different context as in our case we use saliency as attention map and train the network for a new task.


\minisection{Fine-grained recognition:} 
Most of the state of the art general object classification approaches~\cite{wang2017residual,krizhevsky2012imagenet} have difficulties in the fine-grained recognition task, which is more challenging due to the fact that basic-level categories (e.g. different bird species or flowers) share similar shape and visual appearance. 
One reason for this could be attributed to the popular codebook-based image representation, often resulting in the loss of subtle image information that is critical for the fine-grained task. 
%

Most current fine-grained recognition approaches operate on a two-stage pipeline, first detecting some object parts and then categorizing the objects using this information. 
The work of Huang et al.~\cite{huang2016part} first localizes a set of part keypoints, and then simultaneously processes part and object information to obtain highly descriptive representations. 
Mask-CNN~\cite{Wei2018maskcnn} also aggregates descriptors for parts and objects simultaneously, but using pixel-level masks instead of keypoints. 
The main drawback of these models is the need of human annotation for the semantic parts in terms of keypoints or bounding boxes.

%
For this reason, several recent approaches perform fine-grained recognition without explicit part annotations.
Some have attempted to detect semantic parts using co-segmentation, like in ~\cite{krause2015fine}.
In \cite{xie2017lg}, their framework first performs unsupervised part candidates discovery and global object discovery which are subsequently fed into a two-stream CNN in order to model jointly both the local and global features. Alternatively, ~\cite{gosselin2014revisiting} uses Fisher vectors for image representation and shows that larger codebooks are able to model subtle visual details without explicitly modeling parts, which leads to better classification accuracy compared to small codebooks.
Regardless, most fine-grained approaches use the object ground-truth bounding box at test time, achieving a significantly lower performance when this information is not available.
Moreover, automatically discovering discriminative parts might require large amounts of training images.
Our approach is more general, as it only requires image level annotations at training time and could easily generalize to other recognition tasks.


\minisection{Few-shot learning:}  Few-shot learning aims to create models for which very few labeled samples are available. 
Early work on this topic is attributed to Fei-Fei et al.~\cite{Fei_Fei2006}, who showed that, taking advantage of previously learned categories, it is possible to learn new categories using one or very few samples per class. 
More recently, \cite{KRISHNAN20151302} proposed a conditional distance measure that takes into account how a particular appearance model varies with respect to every other model in a model database. The approach has been applied to one-shot gesture recognition.
%
Nowadays, several deep learning-based approaches have emerged to address the problem of few-shot learning. 
We can identify three main strategies. 
One family is based on metric learning. 
In~\cite{vinyals2017match}, the authors propose a framework that trains a network to map a small labeled support set and an unlabeled example to its label. 
An extension of this idea is presented in Prototypical networks~\cite{Prototypica2017}, but in this case each class in the support set has been substituted by a `prototype' (computed as the mean of the samples in the corresponding class), to which each sample is compared.

A second family of approaches in based on meta-learning, i.e. learning a model that given a few training examples of a new task tries to quickly learn a learner model that solves this new task \cite{munk2017meta}. In~\cite{ravi2017optimization}, the authors propose an LSTM-based meta-learner that is trained to optimize a neural network classifier. The meta-learner captures both short-term knowledge within a task and long-term knowledge common among all the tasks. 

Finally, the third family of approaches is based on data augmentation for data-starved classes. In \cite{girshick2017hallucinating}, the authors propose a way to increase (``hallucinate") the number of samples for the classes with limited data. Their method is based on the intuition that certain aspects of intra-class variation generalize across categories, like for instance pose transformations. In practice, for data-rich classes, they use a neural network to learn transformations between pairs of samples and this transformation is later on applied on the real samples from data-starved classes to generate synthetic ones, thus increasing the population of these classes. For the same purpose (i.e. data augmentation for data-starved classes), in \cite{vasconcelos2017aga} the authors propose an attributed-guided augmentation approach
which learns a mapping that allows the creation of  synthetic data by manipulating certain attributes of real data. Thus, the newly created data presents attributes based on user-defined criteria (values). 
Instead of performing the data augmentation in image space, they perform it in feature space. This idea is further extended in \cite{vasconcelos2018fatten}, where the authors use a deep encoder-decoder architecture to generate feature trajectories by exploiting the pose manifold in terms of pose and appearance. 

\section{Saliency Modulation for Scarce Data Object Classification}
\label{sec:method}

Image classification results have improved much since the advent of deep convolutional neural networks~\cite{krizhevsky2012imagenet,Resnet50} due to the excellent visual representations learned by these models. 
Given the great number of parameters of these networks, we require large datasets of labeled data to effectively train them.
For example the popular ImageNet dataset has over 1M labeled images~\cite{russakovsky2015imagenet}. 
Once learned, these strong image representations can be transfered to other related tasks by a process called finetuning. This process allows to use deep learning on tasks for which significantly less labeled data is available. 
In some cases, however, the available data for the target task is so scarce that is still insufficient to finetune large networks and obtain satisfactory results.

\begin{figure}
\centering
\includegraphics[width=\columnwidth]{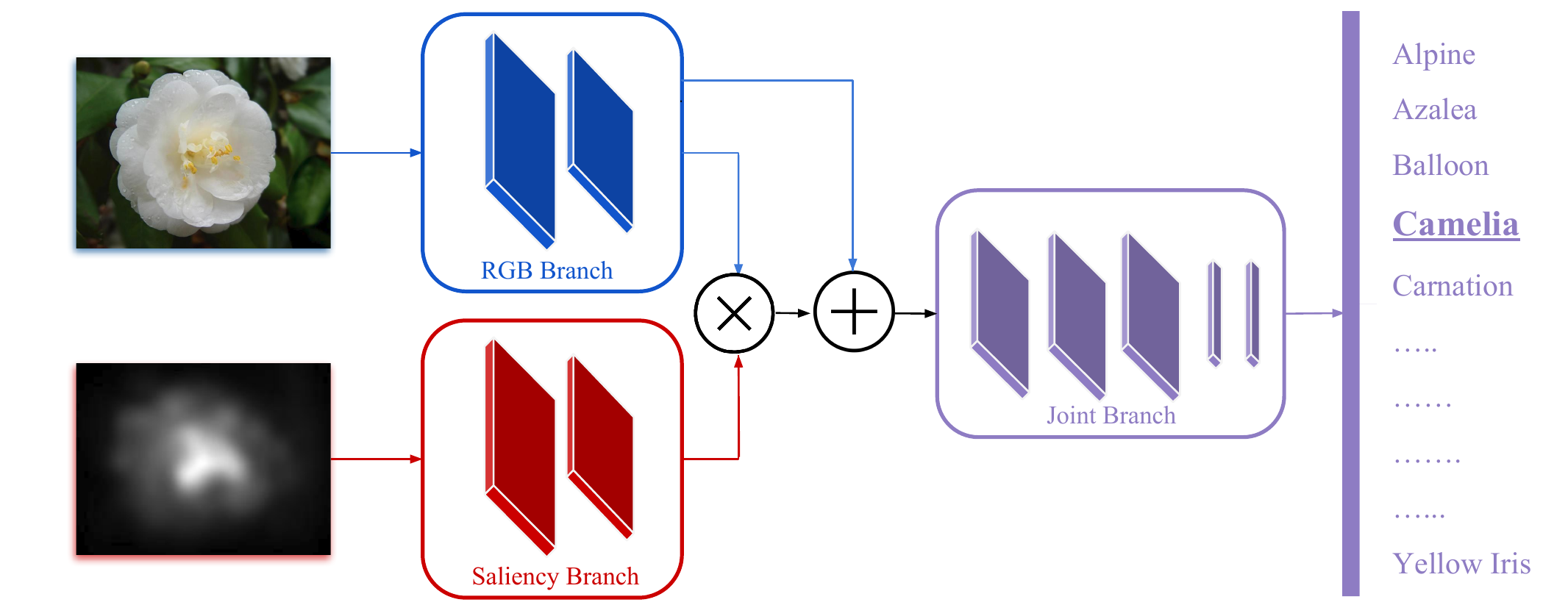}
\caption{Overview of our fine-grained recognition model using saliency information.   
  We process the two  inputs, RGB and Saliency map, through two convolutional layers and then fuse the resulting features with a modulation layer. We then continue processing the fused features with three more convolutional layers and three fully connected layers, ending with the final classification layer.}
\label{fig:overview}
\end{figure}

Saliency is an attentional mechanism which allows humans to focus their limited resources to the most relevant information in the image. Since processing resources are limited, the data is processed in a serial manner, prioritizing those parts that are expected to have high information content. 
In this paper, we investigate another potential application of saliency, namely its function to facilitate the fast learning of new objects in the context of deep neural networks. 
Especially when only a few training examples are available, focusing on the relevant parts of the image could significantly improve the speed of learning, understanding speed as the number of example images required to learn a new class. 
Therefore, we seek to incorporate saliency estimation into an image classification pipeline, with the aim to decrease the data requirements for learning object categories. 


Fig.~\ref{fig:overview} provides an overview of the proposed network architecture. 
Our network contains two branches: one to process the RGB images and one to process their corresponding saliency images, which are pre-computed and given as input. 
They are combined with a \emph{modulation layer} ($\times$ symbol) and further processed by several shared layers of the joint branch to finally end on a classification layer. 
Note how the RGB branch followed by the joint branch correspond to a standard image classification network. 
The novelty of our architecture is the introduction of the saliency branch, which transforms the saliency image to the \emph{modulation image}. 
This modulation image is then used to modulate the features of the RGB branch, putting more emphasis on those features that are considered important for the fine-grained recognition task. 
In the following sections we provide the details of the network architecture, the functioning of the modulation layer, and the saliency methods used. 
We explain our model using AlexNet~\cite{krizhevsky2012imagenet} as base classification network, but the theory could be applied to most convolutional neural network architectures.
We also consider ResNet-50 and ResNet-152~\cite{Resnet50} as base networks in our experiments (sec.~\ref{sec:results}).

\subsection{Combining RGB with Saliency for Image Classification}

Consider a saliency map $s(x,y)$ where $x$ and $y$ are the spatial coordinates. We will assume that saliency maps are of the same size as the original image $I(x,y,z)$, where $z=\{1,2,3\}$ indicate the three color channels of the image. 
A straightforward way to incorporate the saliency into the image classification network is by concatenating the image and the saliency map into an image with four channels such that $I(x,y,4) = s(x,y)$. This strategy has been previously used by Murabito et al.~\cite{murabito2018top} in a classification pipeline that combines two CNN networks: one to compute top-down saliency maps from an RGB image, and a second network that appends the generated saliency map to the RGB image channels to perform image classification.
%
In this case, the classification network only needs to train from scratch the weights of the first layer, the following layers can be initialized with a pretrained network. We call this approach \emph{early fusion} of saliency and image content.

In this article we propose \emph{delayed fusion} of saliency and image content, where we use the saliency map to modulate the features of an intermediate network layer. Consider the output of the $i^{th}$ layer of the network, $l^i$, with dimension $w_i \times h_i \times z_i$. Then we define the modulation with a function $\hat s(x,y)$ as
\begin{equation}
\hat l^i \left( {x,y,z} \right) = l^i \left( {x,y,z} \right) \cdot \hat s\left( {x,y}\right),  \label{eq:forward}
\end{equation}
yielding the saliency-modulated layer $\hat l^i$. 
Here the modulation image $\hat s$ is the output of the saliency branch, which takes $s$ as input (as depicted in Fig.~\ref{fig:overview}). 
Note that we consider a single saliency map $\hat s$ that is independent of the number of feature maps. 
To ensure that  $\hat s$ has the same spatial dimensions as $l^i$, we use a similar architecture for both the saliency branch and the RGB branch.
Concretely, the main difference resides in the size of the channel dimension: the saliency branch takes an intensity image as input (instead of a 3-channel RGB image) and outputs a scalar modulation image of $w_i \times h_i \times 1$ (instead of a $w_i \times h_i \times c_i$ feature map). 
Moreover, we use a sigmoid activation function at the end of the saliency branch, as opposed to the ReLU non-linearity of the RGB branch. 
This ensures that $0 \le \hat s\left( {x,y} \right) \le 1$ and thus provides a suitable range for feature modulation.

In the original architecture, max pooling is performed right after the second convolutional layer. 
In our model, we postpone this max pooling to after the features from both branches are fused, i.e. we perform max pooling on the salience-modulated layer $\hat l^i$.
The reasoning behind this choice is to leverage the greater modulation potential of higher resolution saliency features.
We experimentally show (sec.~\ref{sec:results}) that this results in a small performance boost.

In addition to the formulation in Eq.~\eqref{eq:forward} we also introduce a skip connection from the RGB branch to the beginning of the joint branch, defined as
\begin{equation}
\hat l^i \left( {x,y,z} \right) = l^i \left( {x,y,z} \right) \cdot \left( \hat s\left( {x,y} \right) + 1 \right)\label{eq:forward2}.
\end{equation}
This skip connection is depicted in Fig.~\ref{fig:overview} (+ symbol). 
It prevents the modulation layer from completely ignoring the features from the RGB branch.
This is inspired by a previous work~\cite{Zagoruyko2017} that found this approach beneficial when using attention for network compression. 
We confirm the usefulness of the skip connection in the experiments section, sec.~\ref{sec:results}.

We train our architecture in an end-to-end manner.
The backpropagated gradient from the modulation layer into the image classification branch is equal to
\begin{equation}
\frac{{\partial L}}{{\partial l^i }} = \frac{{\partial L}}{{\partial \hat l^i }} \cdot \left( s +1 \right),
\end{equation}
where $L$ is the loss function of the network. This shows that the saliency map not only modulates the forward pass (see Eq.~\eqref{eq:forward2}), but it also modulates the backward pass in exactly the same manner; in both cases putting more weight on the features that are on locations with high saliency, and putting less weight on the irrelevant features in the background on which the network could potentially overfit. 



\subsection{Training the Saliency Branch}
%
The aim of the saliency branch is to process the saliency map $s(x,y)$ into effective modulation features $\hat{s}(x,y)$ that increase the classification performance when training with scarce data.
The main intuition is that the saliency features $\hat{s}$ will focus the backpropagated gradient to the relevant image features, thereby reducing the required data necessary to train the network. 
The additional saliency branch necessary to compute $\hat{s}(x,y)$ has its own set of parameters and could, in principle, increase the possibility of overfitting. 
We therefore consider two different scenarios to initialize this branch. 
In both cases, we start with an equivalent architecture to the one depicted in Fig.~\ref{fig:overview} but without the saliency branch.
We pretrain this network for image classification on ImageNet~\cite{russakovsky2015imagenet}. Then, we add the saliency branch and apply either of the following options:

\begin{itemize}
\item \emph{Initialization from scratch}: the weights of the saliency branch are randomly initialized using the Xavier method. 

\item \emph{Initialization from pretrained}: the weights of the saliency branch are pretrained on an image classification network for which abundant training data is available.
To do this, we first generate saliency images for the ImageNet validation dataset, which consists of 50K images (40K for training and 10K for validation) using the saliency method of choice. On this dataset we train our method, initializing the saliency branch from scratch. We now have a good pretrained model for the saliency branch too. 
Finally, we use this pretrained network (using both the saliency and RGB branch) to initialize all the weights of our network except the top classification layer.
\end{itemize}

%

\subsection{Saliency input}
The input to the saliency branch is a saliency map. Among the many saliency methods that provide satisfactory results~\cite{mit-saliency-benchmark}, we perform most of our experiments using two of the top performing methods:
\begin{itemize} 
\item iSEEL~\cite{tavakoli2017exploiting} leverages the inter-image similarities to train an ensemble of extreme learners.
The predicted saliency of the input image is then calculated as the ensemble's mean saliency value. Their approach is based on two aspects: (i) the contextual information of the scene and (ii) the influence of scene memorability (in terms of eye movement patterns by resemblance with past experiences).
We use MATLAB code released by the authors.
\item SALICON~\cite{huang2015salicon} exploits the power of high-level semantics encoded in a CNN pretrained on ImageNet. Their approach represents a breakthrough in saliency prediction, by reducing the semantic gap between the computational model and the human perception. 
Their method has two key elements: (i) an objective function based on saliency evaluation metrics and (ii) integration of information at different image scales. We use the open source implementation provided by~\cite{christopherleethomas2016}.
\end{itemize}

Besides these two methods, we also perform experiments with three other approaches for a more comprehensive comparison. 
\begin{itemize}
\item Itti and Koch~\cite{itti1998model}: First, we consider the classical saliency model of Itti et al. Several activation maps, corresponding to multiscale image features (color, intensity and orientations) are generated from the visual input and combined into a single topographical saliency map. A neural network is used to select the most salient locations in order of decreasing magnitude, which could be subsequently analyzed by more complex, higher cognitive level processes. 

\item GBVS~\cite{perona2006gbvs}: The Graph-based Visual Saliency (GBVS) is also a biologically-plausible bottom-up model following the approach proposed earlier by Itti et al., but improving the performance of the generation of activation maps and the normalization/combination step. 
They used the Markovian formalism to describe the dissimilarity and concentration of salient locations of the image seen as a graph.

\item BMS~\cite{sclaroff2013bms}:  Boolean Map based Saliency (BMS) approach computes saliency by analyzing the topological structure of the Boolean maps. These maps are generated by randomly thresholding the color channels. As topological element they choose `sorroundedness' because it better characterizes the image/background segregation.

\end{itemize}

\begin{figure*}[t]
\centering
\includegraphics[width=12.0cm]{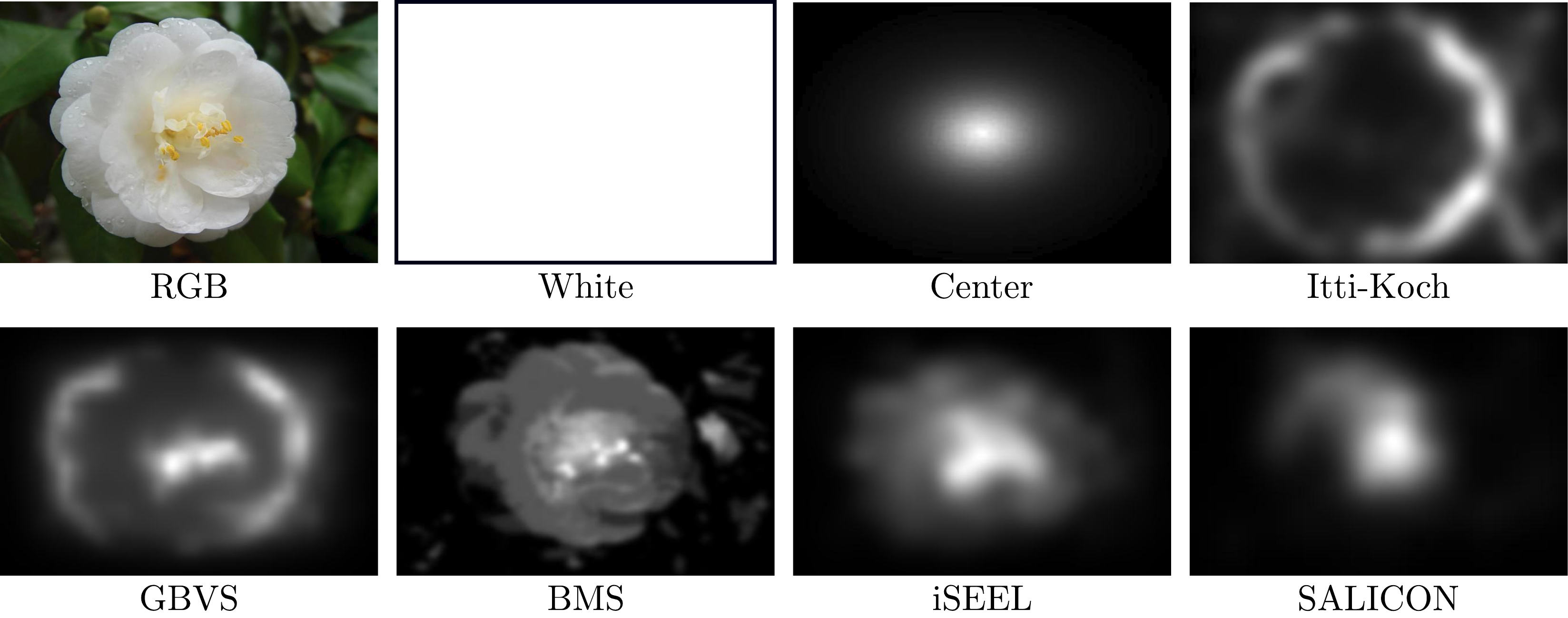}
\caption{Saliency images generated with the different saliency estimation approaches considered, as well as the two baseline saliency maps evaluated, White and Center. We also include the original RGB image for reference.}
\label{fig:saliency_methods}
\end{figure*}

Figure~\ref{fig:saliency_methods} depicts the estimated saliency maps for an example image using the five different saliency methods presented above. 
In addition to these methods, we consider two additional saliency map baselines.
\emph{White} regards all image pixels as equally salient, and thus the saliency maps are uniformly white.
On the other hand, \emph{Center} emulates a center prior by representing saliency as a centered 2-dimensional Gaussian distribution.
These two baselines allow us to determine whether our model is actually leveraging the saliency information contained in the maps, or it is simply adding a general image bias that is beneficial for recognition (e.g. center bias).
We are especially interested in assessing whether saliency methods that obtain higher performance on saliency benchmarks also yield better performance when incorporated into our saliency pipeline.

\section{Experiments}
\label{sec:experiments}

\subsection{Experimental Setup}
\label{sec:setup}




\minisection{Datasets.}
We have performed the evaluation of our approach on four standard datasets used for fine-grained classification
\begin{itemize}
\item  \textit{Flowers}: Oxford Flower 102 dataset~\cite{nilsback2008automated} consists of 8189 images of flowers grouped in 102 classes. Each class contains between 40 and 258 images.

\item \textit{Birds}: is a dataset consisting of 11,788 images of bird species divided in 200 categories~\cite{WelinderEtal2010}. Each image is annotated with its bounding box and the image coordinates of 15 keypoints. However, in our experiments we used the whole image.

\item \textit{Cars}: the dataset in~\cite{krause2013} contains 16,185 images of 196 classes of cars. The data is split into 8,144 training images and 8,041 testing images, where each class has been separated roughly in a 50-50 split.

\item \textit{Dogs}: Stanford Dogs~\cite{KhoslaYaoJayadevaprakashFeiFei_FGVC2011} consists of 20,580 images of different breeds of dogs from around the world grouped in 120 categories. Since some of these images appear also in Imagenet, in our experiments we have discarded the repeated ones.

\end{itemize}

\minisection{Networks.}
Our base network is AlexNet~\cite{krizhevsky2012imagenet}, which consists of five convolutional layers followed by three fully connected layers.
We used the pretrained network on ImageNet~\cite{russakovsky2015imagenet} and fine-tuned it for fine-grained recognition on each dataset for 70 epochs with a learning rate of 0.01 and a weight decay of 0.003.
The top classification layer is randomly initialized using Xavier.
We have attached a saliency branch to this network as shown in Figure ~\ref{fig:overview}. 

For some experiments we have also used the ResNet-50 and ResNet-152~\cite{Resnet50}, consisting of 50 and 152 convolutional layers, respectively, organized in 5 residual blocks. The structure of the saliency branch has been kept the same as in Figure~\ref{fig:overview}, i.e. consisting of two convolutional layers and having a ReLu activation function after the first one and a sigmoid function after the second.

\minisection{Evaluation protocol.}
For all the above datasets, we randomly select and fix 5 images for test, 5 for validation, and keep the rest for training. 
We do this for each class in the dataset independently. In order to investigate different data scarcity levels, 
we train each model with subsets of $k$ training images for $k\in\{1,2,3,5,10,15,20,25,30,K\}$, where $K$ is the total number of training images for the class, which does not include the 10 held out images for validation and test. Contrarily to current few-shot approaches, this setting grants us a more complete disclosure of the results of our model under multiple limited-data scenarios.
We use accuracy in terms of percentage of correctly classified samples as evaluation measure.
We train and test each model five times with different random initializations, and show the average performance for the five runs.

\subsection{Experimental Results}
\label{sec:results}

In the experimental section we evaluate the best strategies for fusing the saliency and RGB branches, compare several network architectures, evaluate various saliency methods as input to the saliency branch, and compare our results with state of the art on standard benchmark datasets for fine-grained object recognition.



\minisection{Optimal architecture.} 
In order to justify the design choices in our model, we present here multiple architectural variations to integrate saliency information into a neural network.
We call \emph{Baseline-RGB} to the original network model, which only contains the RGB branch and thus does not use any saliency information.
We test an \emph{Early fusion} model in which the saliency image is directly concatenated to the RGB input.

We consider several variants of our model in which delayed fusion is performed at different network levels, indicated as Fusion L1 for fusion after layer 1 (similarly for Fusion L2, L3, L4, and L5). In all cases, we use a two-layer saliency branch, indicated by S2.  Moreover, we evaluate whether performing the fusion after the pooling layer is a better option than doing it before.
Finally, we include a model without the skip connection from the RGB branch to the joint branch (No SC).
\begin{table*}[t]
\centering
\resizebox{\textwidth}{!}{%
        \begin{tabular}{l|ccccccccccc}
            \hline
            Method & 1 & 2 & 3 & 5 & 10 & 15 & 20 & 25 & 30 & $K$ & $\textbf{AVG}$\\ \hline 
            Baseline-RGB & 31.8 & 45.8 & 53.1 & 63.6 & 72.4 & 76.9 & 81.2 & 85.1 & 87.2 & 88.0 & 68.5\\ \hline
            Early Fusion &19.3 &25.7 & 30.1& 40.8& 60.9&69.2 &75.3 &79.9 &82.4 & 83.7 & 56.7\\ \hline
            Fusion L1-S2 &33.3 &47.9&54.3 & 65.1&71.9 &76.3 &82.1 &85.9 &87.9 &90.7 &69.5 \\ 
            Fusion L2-S2 & \textbf{34.7}& \textbf{49.3}&55.2 & 65.2& \textbf{72.7}& 76.7& \textbf{83.9}& \textbf{86.5}& \textbf{89.1}&\textbf{91.3} & \textbf{70.5}\\
            Fusion L3-S2  &32.9 & 46.7&54.1 &64.9 &71.7 &74.4 &82.3 &85.1 &87.3 & 89.1 & 68.9\\  
            Fusion L4-S2  &32.5 & 48.9&54.0 &65.1 &71.7 &73.5 &81.0 &84.9 &87.2 &88.8 & 68.2 \\
            Fusion L5-S2  &32.5 & 48.9&54.0 &63.3 &71.1 &73.3 &81.0 &84.3 &87.2 &88.7 & 68.4  \\\hline 
            
            Fusion L2-S2 + After pool & 34.3&49.1& \textbf{55.5}&\textbf{66.0}&72.1&\textbf{77.5}&83.6&85.6&88.9& 90.2 & 70.2 \\ \hline 
            Fusion L2-S2 + No SC &33.9 &48.1 &55.1 &65.1 &71.1 &77.6 &82.4 &86.3 &88.1&90.9 &69.9 \\ \hline 

        \end{tabular}
}       
        \caption{Results for the baseline model and different variations of our architecture incorporating saliency information.The results correspond to the classification accuracy on the Flowers dataset~\cite{nilsback2008automated} with AlexNet~\cite{krizhevsky2012imagenet}. Each column indicates the number of training images used, and the rightmost column shows the average}.
        \label{tab:architecture2}
\end{table*}



We evaluate all models on \emph{Flowers}~\cite{nilsback2008automated} with AlexNet~\cite{krizhevsky2012imagenet} and using iSEEL~\cite{tavakoli2017exploiting} as the saliency method of choice.
Table~\ref{tab:architecture2} shows the results for different number of training images. 
First, we observe how the performance of all methods steadily grows when increasing the number of training images.
In general, incorporating saliency information helps when fused within the network, but damages the accuracy if concatenated to the input image.
We attribute this to the need to learn a low-level filter from scratch, which in turn affects the feature representation at higher levels.
Performing the fusion immediately after the second convolutional layer seems to be the best option.
Fusing before or after the pooling layer leads to similar results, the advantage of fusing higher resolution saliency features gives only a marginal boost.
Finally, the skip connection from the RGB branch to the joint branch is also beneficial.

We have also explored different architectures for the saliency branch.
We first assess whether an additional convolutional layer in the saliency branch leads to better performance.
Table~\ref{tab:architecture3} presents the comparison between a two-layer saliency branch (S2) and a three-layer version (S3). 
For completeness, we explore merging after the second layer of the RGB branch (L2) as in previous experiments, and merging after the third layer (L3).
We observe how an extra layer does not further improve the model's performance.
Alternatively, we investigate whether having fewer parameters in the saliency branch achieves higher results.
We evaluate with 75\% and 50\% fewer parameters by reducing the number of output channels in the first layer.
Table~\ref{tab:architectureFilters} shows how reducing the number of parameters in the saliency branch slightly reduces the final performance.

\begin{table*}[t]
\centering
\resizebox{\textwidth}{!}{%
        \begin{tabular}{l|ccccccccccc}
            \hline
            Method & 1 & 2 & 3 & 5 & 10 & 15 & 20 & 25 & 30 & $K$ & $\textbf{AVG}$\\ \hline 
            Baseline-RGB & 31.8 & 45.8 & 53.1 & 63.6 & 72.4 & \textbf{76.9} & 81.2 & 85.1 & 87.2 & 88.0 & 68.5\\ \hline
            
            Fusion L2-S2 & \textbf{34.7}& \textbf{49.3}&55.2 &\textbf{ 65.2}& 72.7& 76.7& \textbf{83.9}& \textbf{86.5}& \textbf{89.1}&\textbf{91.3} & \textbf{70.5}\\
            Fusion L3-S2  &32.9 & 46.7&54.1 &64.9 &71.7 &74.4 &82.3 &85.1 &87.3 & 89.1 & 68.9\\  \hline 
            Fusion L2-S3  &34.5 & 48.2&\textbf{55.9 }&65.0 &\textbf{72.8 }&76.1 &83.0 &\textbf{86.5} &89.0 &91.0 &70.2  \\
            Fusion L3-S3  &33.1 & 49.3&54.2 &65.1 &72.1 &74.9 &82.9 &85.3  &88.0 &89.0 &69.4  \\\hline

        \end{tabular}
}       
        \caption{Results on Flowers~\cite{nilsback2008automated} with AlexNet~\cite{krizhevsky2012imagenet} using two (S2) or three (S3) convolutional layers for the saliency branch.}
    \label{tab:architecture3}
\end{table*}

\begin{table*}[t]
\centering
\resizebox{\textwidth}{!}{%
        \begin{tabular}{l|ccccccccccc}
            \hline
            Method & 1 & 2 & 3 & 5 & 10 & 15 & 20 & 25 & 30 & $K$ & $\textbf{AVG}$\\ \hline 
            Baseline-RGB & 31.8 & 45.8 & 53.1 & 63.6 & 72.4 & 76.9 & 81.2 & 85.1 & 87.2 & 88.0 & 68.5\\ \hline
            
            Fusion L2-S2 (100\%) & \textbf{34.7}& \textbf{49.3}&55.2 & 65.2& \textbf{72.7}& 76.7& \textbf{83.9}& \textbf{86.5}& \textbf{89.1}&\textbf{91.3} & \textbf{70.5}\\
            Fusion L2-S2 (75\%)  &34.7 & 49.0& 55.3 & 65.1 &72.0 &77.0 &83.3 &85.9 &88.3 & 89.1 & 70.0 \\  
            Fusion L2-S2 (50\%)  &34.7 &49.1 & \textbf{55.9} & 65.1 &71.8 &\textbf{77.1 }&83.5 &86.2 &88.0&89.0 &70.0  \\\hline

        \end{tabular}
}       
        \caption{Results on Flowers~\cite{nilsback2008automated} with AlexNet~\cite{krizhevsky2012imagenet} when reducing the number of parameters of the saliency branch.}
    \label{tab:architectureFilters}
\end{table*}




\begin{figure}[h!]
\begin{center}
\includegraphics[width=\columnwidth]{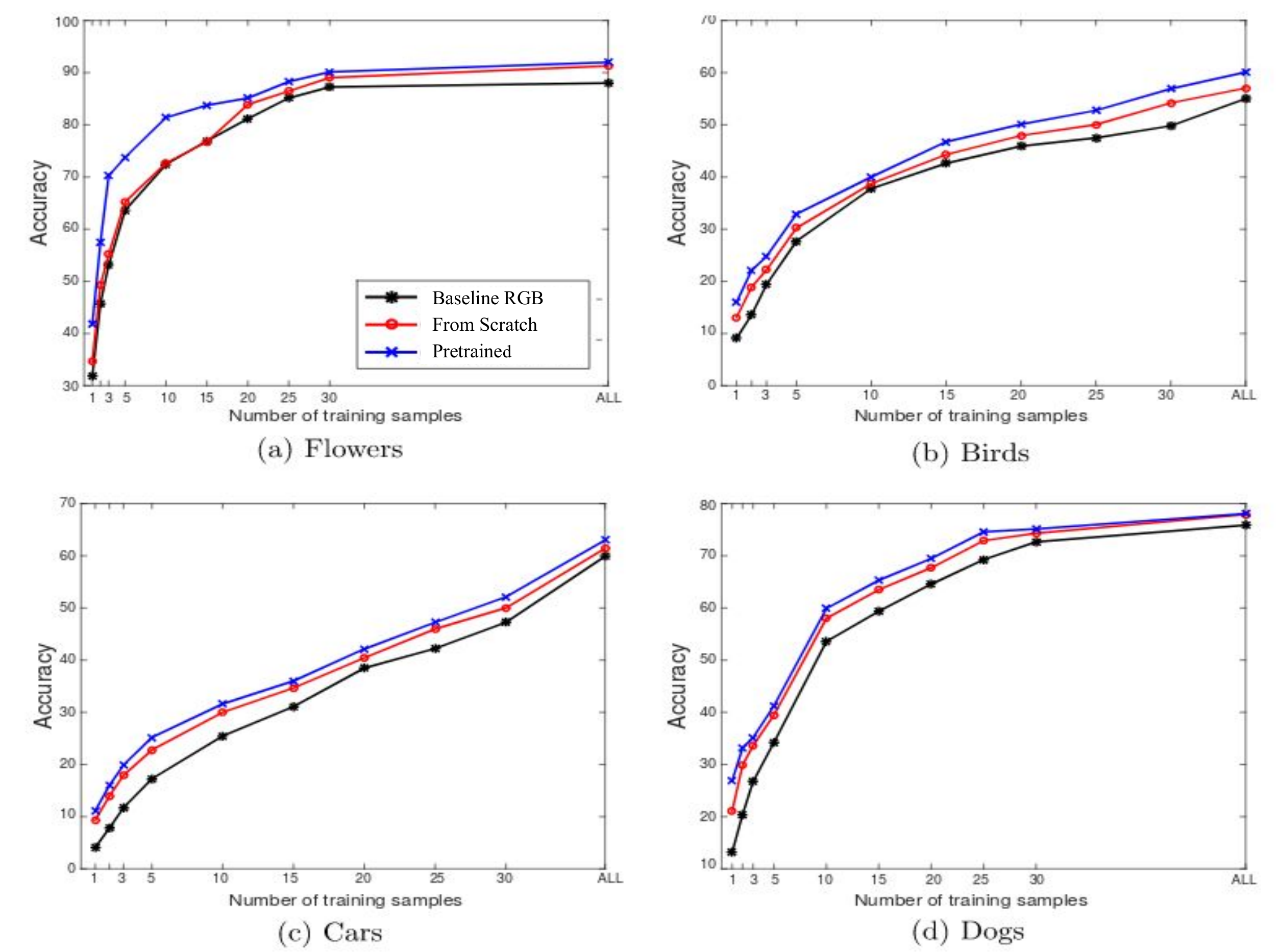} 

\end{center}\caption{Experiments on four datasets using iSEEL~\cite{tavakoli2017exploiting} to generate the saliency maps. \textit{Baseline-RGB} is compared against two different ways to initialize the saliency branch of our model: from scratch and pretrained on ImageNet~\cite{russakovsky2015imagenet}.}
 \label{fig:Plots_Datasets_ISEEL}
\end{figure}

\minisection{Pretraining the saliency branch on ImageNet.}
As described in section~\ref{sec:method}, we consider two alternative ways of initializing the saliency branch: from scratch and pretrained on ImageNet~\cite{russakovsky2015imagenet}.
In this section, we compare these two approaches with respect to the Baseline-RGB. The experiments are performed on \textit{Flowers} dataset (see Figure~\ref{fig:Plots_Datasets_ISEEL}a) and represent the classification accuracy versus the number of training samples.
Adding a saliency branch initialized from scratch already outperforms the baseline using only RGB (see also Tab.~\ref{tab:architecture2}), and pretraining this branch with ImageNet further increases the performance in a systematic and substantial manner.
Our method with pretraining is especially advantageous in the scarce-data domain (i.e $<20$ images per class).
For example, we obtain a better performance than the baseline using half the data, 10 images/class vs. 20 images/class, respectively.
Furthermore, in the very low-range of number of samples we obtain similar performance with only \emph{one third} of the samples (3 images/class vs. 10 images/class).
Finally, our saliency branch is still beneficial even when using all available training samples.
In fact, our method trained with a limited number of samples (around 25 per class) already surpasses the final performance of baseline using all samples.

\begin{figure}[t]
\begin{center}
\includegraphics[width=0.8\columnwidth]{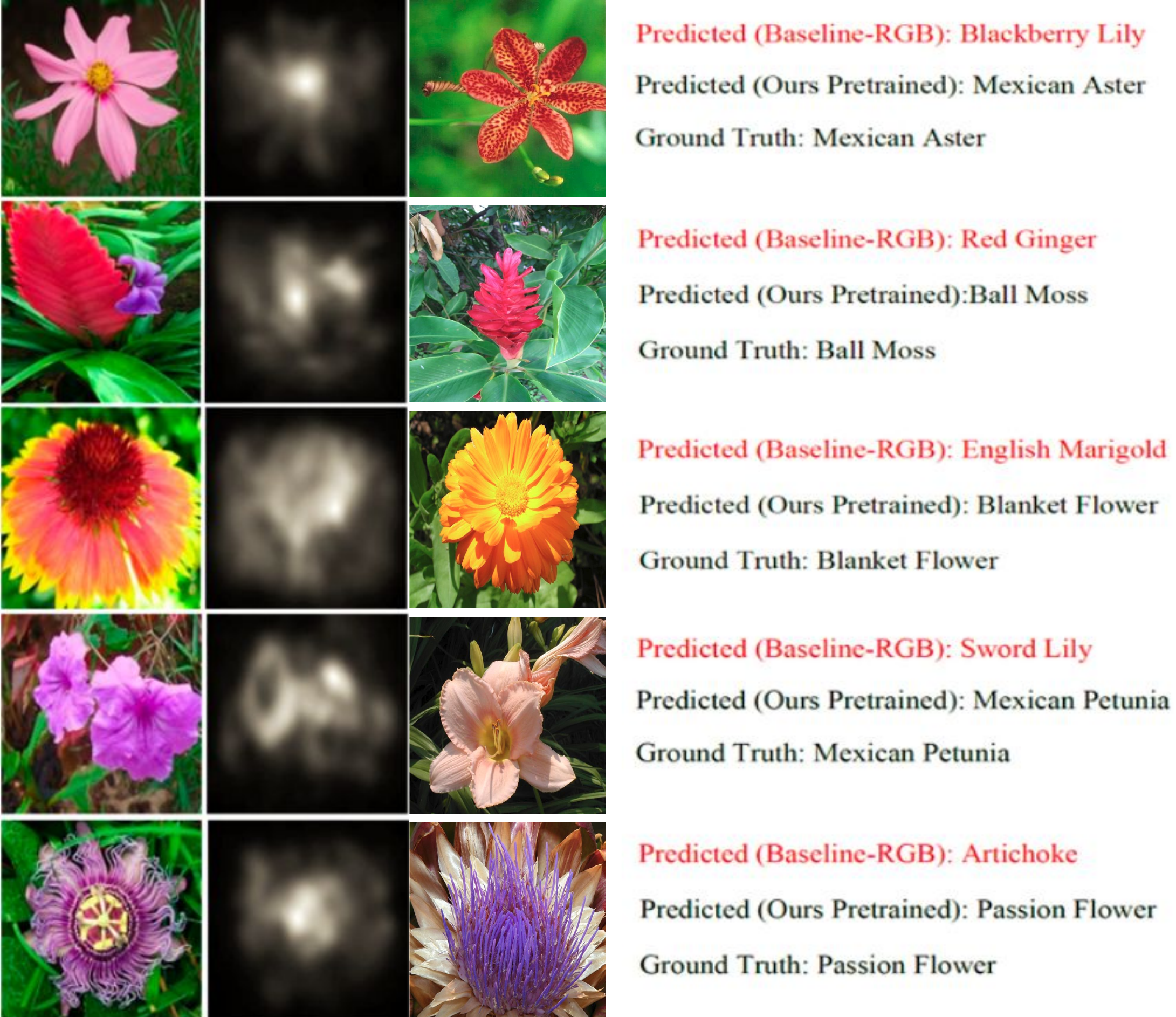} 
\end{center}
\caption{Some success examples on Flowers~\cite{nilsback2008automated}: when the prediction done by Baseline-RGB fails to infer the right label for some test images, but the prediction by our approach is correct. From left to right: input image, saliency images generated with iSEEL~\cite{tavakoli2017exploiting}, example image of the class with which the input image was wrongly predicted.} 
 \label{fig:acierto}
\end{figure}

\begin{figure}[t]
\begin{center}
\includegraphics[width=0.8\columnwidth]{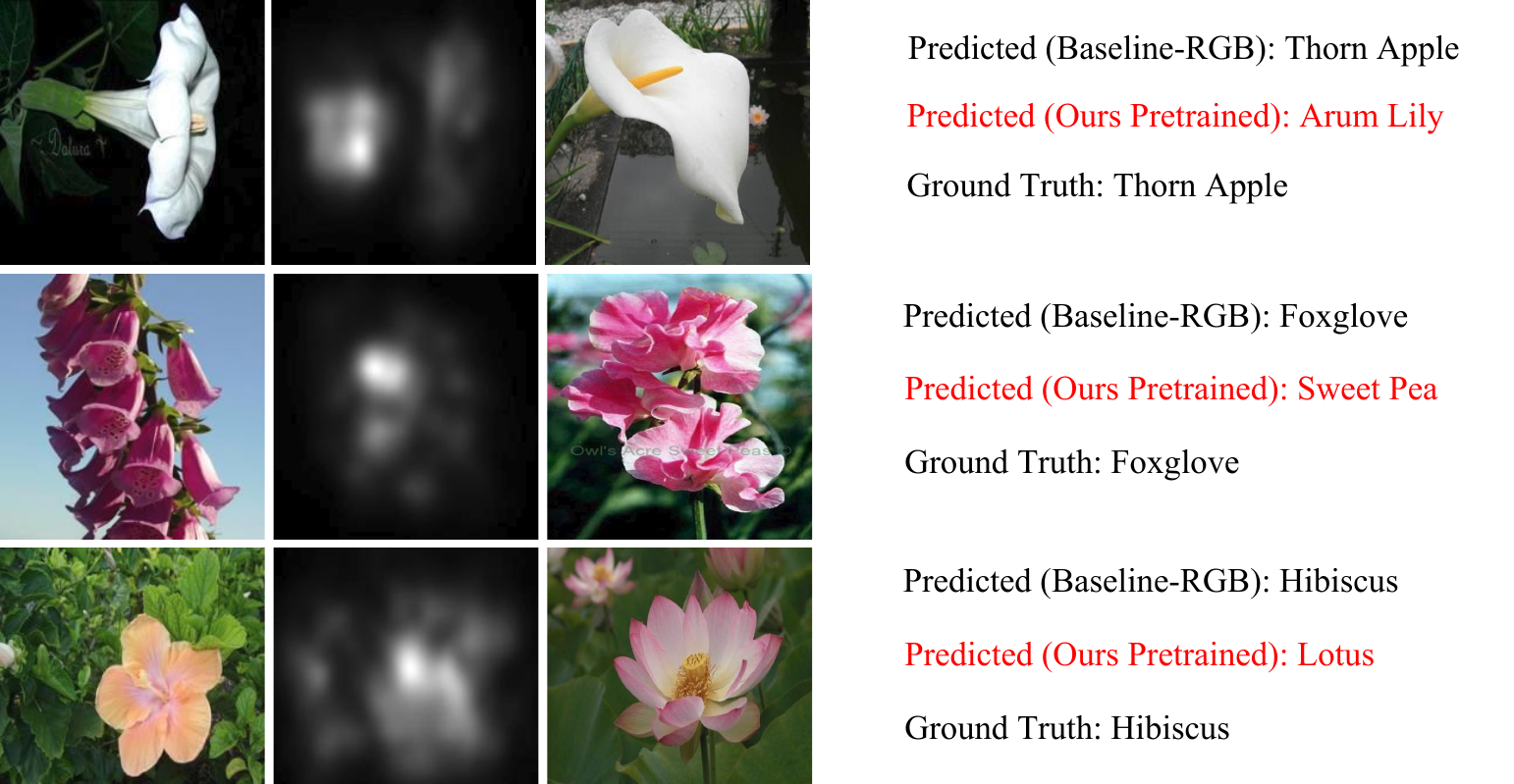} 
\end{center}
\caption{Some failure examples on Flowers~\cite{nilsback2008automated}: when the prediction done by our method fails to infer the right label for some test images, but the prediction by Baseline-RGB is correct. From left to right: input image, saliency images generated with iSEEL~\cite{tavakoli2017exploiting}, example image of the class with which the input image was wrongly predicted.}
 \label{fig:error}
\end{figure}

Figure \ref{fig:acierto} shows some qualitative results for the case when the pretrained version of our approach predicts the correct label, meanwhile the Baseline-RGB fails. Alternatively, figure \ref{fig:error} depicts the opposite case: the Baseline-RGB predicts the correct label of the test images, meanwhile the pretrained version of our approach fails. In both cases, the saliency images have been generated using the iSEEL method. A possible explanation for the failures in this latter case could be that the saliency images are not able to capture the relevant region of the image for fine-grained discrimination. Thus, the salience-modulated layer focuses on the wrong features for the task.

\minisection{Different datasets.}
Besides \textit{Flowers} dataset, we validate our approach on three other datasets: \textit{Birds}, \textit{Cars} and \textit{Dogs} (see figures \ref{fig:Plots_Datasets_ISEEL}b, c, and d, respectively). 
We follow the same experimental protocol as in the \emph{Flowers} case.
We can see how most trends observed in \emph{Flowers} also apply to these datasets. 
For example, incorporating saliency information improves the classification accuracy, especially when data is scarce.
Moreover, pretraining the saliency branch is beneficial for our method and leads to a further performance boost.
Even when using all available samples, our method outperforms the baseline model.
Therefore, we can claim that our approach successfully generalizes to other fine-grained datasets.

\minisection{Confirmation of intuition.} Our method is based on the idea that adding a saliency branch helps the network to focus on the relevant image regions during the training. To verify that this is actually happening we propose the following experiment: we measure if the percentage of backpropagated gradient magnitude which passes through the relevant image regions is increased by our proposed network architecture. We perform this experiment on the Birds dataset for which we have access to bounding box information of the birds (defining the relevant region). We measure the percentage of backpropagated gradient energy which is in the bounding box of the bird (this is computed by dividing the gradient magnitude in the bounding box by the gradient energy in the whole image). We measure this just before the third convolutional layer for AlexNet (which is just before the joint branch in Figure~\ref{fig:overview}), and we measure this for both the network with and without saliency branch. 

\begin{figure}[h!]
\begin{center}
\includegraphics[width=.7\columnwidth]{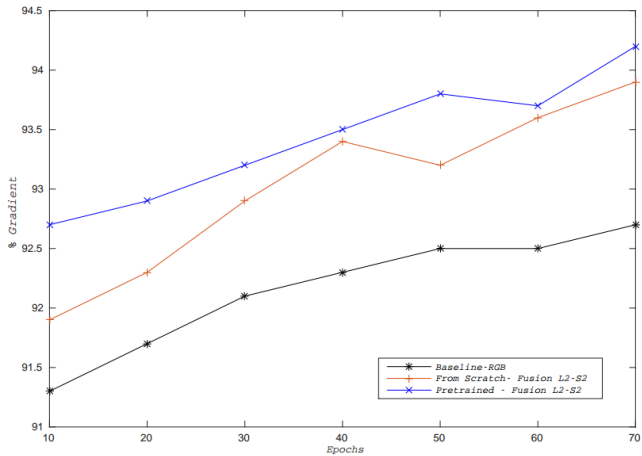} 
\end{center}\caption{Average percentage of the total backpropagated gradient energy per epoch that is inside the bird bounding box. The graph shows that for our approach significantly more backpropagated gradient is on the relevant image region (for both the version trained from scratch and the version with pretrained saliency branch).}
 \label{fig:GradientInside_5Images}
\end{figure}

The results are presented in Figure~\ref{fig:GradientInside_5Images}. The results show that the percentage of backpropagated gradient that passes through the relevant image regions is higher for our approach. As expected it is even higher for the network with the pretrained saliency branch. However the gap with the network trained from scratch diminishes with the number of epochs. The fact that more backpropagated gradient energy goes through the relevant image regions may explain why our method obtains better results than the standard baseline method.

\begin{figure}[h]
\begin{center}
\includegraphics[width=0.8\columnwidth]{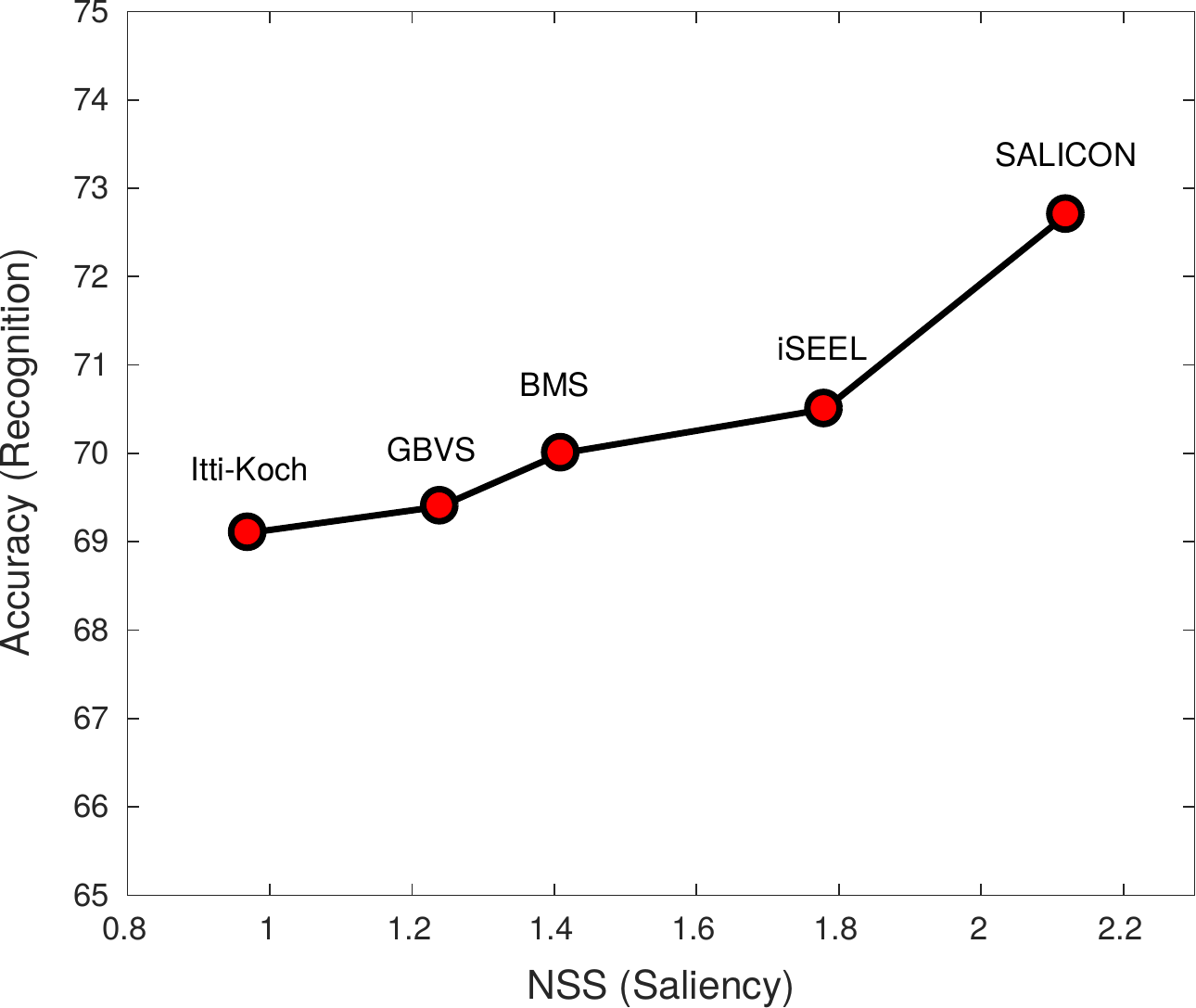} 
\end{center}
\caption{Correlation between the performance of the saliency method in terms of NSS and the fine-grained recognition accuracy of our method using the corresponding saliency model. Results with AlexNet~\cite{krizhevsky2012imagenet} on Flowers~\cite{nilsback2008automated}.}
 \label{fig:correlation}
\end{figure}

\minisection{Different saliency methods.} 
Table~\ref{tab:saliency} presents results on the \textit{Flowers} using our full AlexNet model combined with the different input saliency maps.
We can observe how, instead of helping, the two saliency baselines are actually hurting the method performance with respect to the Baseline-RGB.
We hypothesize that this is due to the noise introduced in the network's internal representation when the input saliency map is independent of the input image.
On the other hand, all the saliency estimation methods increase the method performance, especially in the scarce-data range (i.e. $<10$ images). 
Moreover, better saliency methods (e.g. iSEEL and SALICON) result in higher accuracies.
In order to experimentally confirm this observation, we show in Fig.~\ref{fig:correlation} the accuracy of our image classification model as a function of the saliency estimation performance of the corresponding method.
We measure saliency estimation performance in terms of Normalized Scanpath Saliency (NSS), which is the official measure currently used by the popular MIT saliency benchmark~\cite{mit-saliency-benchmark} to sort all the participating methods.
There is indeed a clear linear correlation, supported quantitatively by a Pearson product-moment correlation coefficient of 0.95.
Therefore, we conclude that our model is agnostic to the saliency method employed. More importantly, it shows that better saliency methods (evaluated based on saliency estimation) actually lead to better image classification performance once integrated into an object recognition pipeline. This observation can be a motivation for saliency research: it not only leads to better saliency estimation but indirectly also contributes to improved object recognition.  

\begin{table*}[tb]
\centering
\resizebox{\textwidth}{!}{%
        \begin{tabular}{l|ccccccccccc}
            \hline
            Method & 1 & 2 & 3 & 5 & 10 & 15 & 20 & 25 & 30 & $K$ & $\textbf{AVG}$\\ \hline 
            Baseline-RGB & 31.8 & 45.8 & 53.1 & 63.6 & 72.4 & 76.9 & 81.2 & 85.1 & 87.2 & 88.0 & 68.5\\ \hline
            Baseline-White & 23.1 & 29.7 &37.2  &55.1  &66.9  &73  &82.5  & 84.8 &86.6  &87.9 & 62.7\\
            Baseline-Center & 24.3 & 30.3 &39.2  &55.7  &68.3  &74.1  &82.7  &84.5  & 86.8 &87.8 & 63.4\\ \hline
            
            Itti-Koch~\cite{itti1998model} &32.8&46.8  & 53.9 & 64.0 & 72.9& 77.1 & 82.9 & 85.4 & 87.1 &88.3 & 69.1\\
            GBVS~\cite{perona2006gbvs} & 33.3 &46.9  &54.0  & 64.1& 73.0 & 77.3 & 83.1&85.7  &87.5  &88.8 & 69.4\\   
  BMS~\cite{sclaroff2013bms} & 34.2 & 47.3 & 54.9 &64.8 & 73.3 & 77.8 & 83.4 & 86.1 & 88.1 &90.1 & 70.0\\           
 iSEEL~\cite{tavakoli2017exploiting} & 34.7& 49.3&55.2 & 65.2& 72.7& 76.7& 83.9& 86.5& 89.1&91.3 & 70.5\\ 
SALICON~\cite{jiang2015salicon} & \textbf{37.6} & \textbf{51.9}& \textbf{57.1} & \textbf{68.5}& \textbf{75.2} & \textbf{79.7}& \textbf{84.9} & \textbf{88.2}& \textbf{91.2} & \textbf{92.4} & $\textbf{72.7}$\\ 
  \hline
        \end{tabular}

}        
        \caption{Comparison of different saliency methods regarding the effect on our model. The results correspond to the classification accuracy on the Flowers dataset~\cite{nilsback2008automated} when using our full model with AlexNet~\cite{krizhevsky2012imagenet} as base network. Each column indicates the number of training images used, and the rightmost column shows the average.}
        \label{tab:saliency}
\end{table*}


\minisection{Different base networks.}
In order to evaluate the generality of our approach across different base networks, we have considered ResNet-50 and ResNet-152 as alternatives to AlexNet. 
We have tested several possible fusion architectures (Tables~\ref{tab:architectureResnet} and~\ref{tab:architectureResnet152}), but the optimal performance has been obtained when the fusion between the RGB and saliency branches takes place after the fourth residual block, with a two-layer saliency branch (Block4-S2). 
Results in Table~\ref{tab:baseNetwork} show the classification accuracy achieved on \textit{Flowers} when using ResNet-50 and ResNet-152 with SALICON salieny maps. Furthermore, we compared our two initialization methods for the saliency branch (from scratch and pretrained on ImageNet) against the Baseline-RGB. Although under both initializations we obtained higher accuracy, the one that performs the best is the pretrained. These results confirm the trend already observed for AlexNet regarding the benefits of pretraining the saliency branch as shown in Fig.~\ref{fig:Plots_Datasets_ISEEL}. 

\begin{table*}[h]
\centering
\resizebox{\textwidth}{!}{%
        \begin{tabular}{l|ccccccccccc}
            \hline
            Method & 1 & 2 & 3 & 5 & 10 & 15 & 20 & 25 & 30 & $K$ & $\textbf{AVG}$\\ \hline 
            Baseline-RGB & 39.1& 59.6 & 67.8 & 81.6& 89.0 &91.7 & 92.7& 93.0& 93.0& 95.4 &80.3  \\ \hline
            Block1-S2     & 38.0& 59.2 & 68.0 & 80.7& 88.8 &91.0 & 91.9& 92.0& 92.1& 94.8 &79.6 \\ 
            Block2-S2     & 38.2& 59.5 & 68.0 & 81.4& 90.0 &91.6 & 92.0& 92.4& 93.0& 94.9 &80.1 \\ 
            Block3-S2     & 39.3& 62.9 & 68.5 & 83.0& 90.0 &92.1 & 93.5& 94.9& 93.4& 95.9 &81.4 \\ 
            Block4-S2     & \textbf{45.8}& \textbf{64.3} & \textbf{72.8} & \textbf{83.0}& \textbf{90.5} &\textbf{93.0} & \textbf{93.9}& \textbf{94.6}& \textbf{93.7}& \textbf{96.7} &\textbf{82.7}\\
            Block5-S2     & 38.2& 57.9 & 65.9 & 80.8& 87.1 &90.9 & 91.1& 91.2& 91.9& 92.6 &78.8\\ 
            \hline
        \end{tabular}
}       
        \caption{Results for the baseline model and different variations of our architecture incorporating saliency information in different blocks. The results correspond to the classification accuracy on the Flowers dataset~\cite{nilsback2008automated} with ResNet-50~\cite{Resnet50}. Each column indicates the number of training images used, and the rightmost column shows the average}.
        \label{tab:architectureResnet}
\end{table*}

\begin{table*}[h!]
\centering
\resizebox{\textwidth}{!}{%
        \begin{tabular}{l|ccccccccccc}
            \hline
            Method & 1 & 2 & 3 & 5 & 10 & 15 & 20 & 25 & 30 & $K$ & $\textbf{AVG}$\\ \hline 
            Baseline-RGB & 39.0 & 60.1 & 68.0 & 82.5 & 89.0 & 92.0 & 92.1 & 93.3 & 94.2 & 95.8 & 80.6  \\ \hline
            Block1-S2    & 39.0 & 59.9 & 68.0 & 82.1 & 88.6 & 91.9 & 92.2 & 93.0 & 94.2 & 95.1 & 80.4  \\ 
            Block2-S2    & 38.8 & 60.2 & 68.2 & 83.0 & 90.2 & 92.2 & 93.0 & 94.0 & 94.0 & 96.2 & 81.0\\ 
            Block3-S2    & \textbf{43.0} & 63.7 & 68.9 & 83.1 & 90.2 & 92.1 & 93.1 & 94.3 & 96.1 & 96.3 & 82.1\\ 
            Block4-S2    & 42.6 & \textbf{64.2} & \textbf{70.9} & \textbf{85.5} & \textbf{90.9} & \textbf{92.7} & \textbf{94.0} & \textbf{95.0} & \textbf{97.0} & \textbf{97.8} & \textbf{83.1}\\
            Block5-S2    & 39.0 & 58.0 & 65.8 & 80.3 & 87.1 & 90.8 & 91.5 & 92.0 & 92.3 & 92.7 & 79.0\\ \hline
        \end{tabular}
}       
\caption{Results for the baseline model and different variations of our architecture incorporating saliency information in different blocks. The results correspond to the classification accuracy on the Flowers dataset~\cite{nilsback2008automated} with ResNet-152~\cite{Resnet50}. Each column indicates the number of training images used, and the rightmost column shows the average}.
        \label{tab:architectureResnet152}
\end{table*}

\minisection{Comparison with standard dataset splits.}
All previous experiments use a custom data split consisting of a fixed test set of 5 images and a varying number of training images. In order to enable comparisons with published results by other methods, we perform here experiments using the standard data split of each dataset, employing the entirety of the corresponding given sets for training and evaluation.
Table~\ref{tab:state_art} presents results for our approach and several state of the art fine-grained recognition approaches for Flowers, Birds, and Cars datasets. 
We discard Dogs dataset due to the overlap with the ImageNet images already used for pretraining the network, as they can no longer be ignored when using the full sets.
Our approach uses SALICON saliency and ResNet152 as base network, which is equivalent to the networks used by the most recent works. 
Our method is competitive with specialized fine-grained approaches, despite the more sophisticated techniques included in those (e.g. part localization), some of which might be complementary to our saliency modulation.
Moreover, our approach is especially beneficial in the scarce training data regime, whereas some of the state of the art methods may not work under these conditions.

\begin{table*}[t]
\centering
\resizebox{\textwidth}{!}{%
        \begin{tabular}{l|ccccccccccc}
            \hline
           Method & 1 & 2 & 3 & 5 & 10 & 15 & 20 & 25 & 30 & $K$ & $\textbf{AVG}$\\ \hline 
            Baseline-RGB Resnet-50 & 39.1 & 59.6& 67.8 &81.6  & 89.0 & 91.7 & 92.7&93.0  & 93.0 & 95.4& 80.3 \\ 
            Resnet-50 Block4-S2 From Scratch & 45.8 & 64.1&71.8 & 83.0 & 90.5 & 93.0 & 93.9&  94.6& 93.7 & 96.7& 82.7\\ 
            Resnet-50 Block4-S2 Pretrained &\textbf{47.1} & 65.2  & 72.9 &83.8  &91.3 & 93.9 &94.6 &95.4  &94.7  & 97.4& 83.6\\ 
           \hline 
           Baseline-RGB Resnet-152 & 39.0 & 60.1 & 68.0 & 82.5 & 89.0 & 92.0 & 92.1 & 93.3 & 94.2 & 95.8 & 80.6  \\
           Resnet-152  Block4-S2 From Scratch & 42.6 & 64.2 & 70.9 & 85.5 & 90.9 & 92.7 & 94.0 & 95.0 & 97.0 & 97.8 & 83.1 \\
           Resnet-152 Block4-S2 Pretrained & 46.9 & \textbf{65.5} & \textbf{73.0} & \textbf{84.7} & \textbf{92.0} & \textbf{94.2} & \textbf{95.3} & \textbf{95.8} & \textbf{97.3} & \textbf{98.1} & \textbf{84.3}   \\ \hline
        \end{tabular}
}        
        \caption{Results on Flowers~\cite{nilsback2008automated} using ResNet-50 and Resnet-152~\cite{Resnet50} as base networks and SALICON~\cite{jiang2015salicon} as saliency method.}
        \label{tab:baseNetwork}
\end{table*}



    

\begin{table*}[h!]
\centering
\resizebox{0.85\textwidth}{!}{%
        \begin{tabular}{l|ccc}
            \hline
             Method & Flowers & Birds & Cars  \\ \hline 
            Krause et al. ~\cite{krause2015fine} & - & 82.0 & 92.6 \\
            RA-CNN ~\cite{RA_CNN2017}           & - & 85.3 & 92.5 \\
            Bilinear-CNN ~\cite{Bilinear_CNN}         & - & 84.1 & 91.3 \\
            Compact Bilinear Pooling ~\cite{Compact_Bilinear_Yang} & - & 84.3 & 91.2 \\
            Low-rank Bilinear Pooling ~\cite{Low_rank_Kong} & - & 84.2 & 90.9 \\
            Cui et al. (with Imagenet) ~\cite{Cui_large2018} &96.3 & 82.8& 91.3\\
            MA-CNN  ~\cite{MA_CNN} & - & \textbf{86.5} & 92.8 \\
            Ge-Yu ~\cite{GE_YU2017} & 90.3 & - & - \\
            DLA  ~\cite{DLA2017} & - & 85.1 & \textbf{94.1} \\
            
            \hline
            Ours (Resnet152 Block4-S2 From Scratch) & 96.4 &85.6 & 92.1 \\
            Ours (Resnet152 Block4-S2 Pretrained) & \textbf{97.8}& 86.1 & 92.4\\ \hline
       
        \end{tabular}

}        
        \caption{Comparison with state of the art methods for domain-specific fine-grained recognition using the standard data splits of Flowers~\cite{nilsback2008automated}, Birds~\cite{WelinderEtal2010} and Cars~\cite{krause2013}.
        Our approach uses ResNet-152~\cite{Resnet50} as base network and SALICON~\cite{jiang2015salicon} saliency maps.}
        \label{tab:state_art}
\end{table*}

\minisection{Comparison with few-shot method.}
\begin{table}
\centering
 \resizebox{0.7\textwidth}{!}{%
        \begin{tabular}{c|c|c}
            \hline
            Method & 20-way 5-shot & 102-way 5-shot\\ \hline 
            Prototypical networks~\cite{Prototypica2017} & 53.8 & 26.2 \\
            Ours & 81.0 & 73.8 \\
  	\hline

        \end{tabular}
        }
        \caption{Results for few-shot classification on Flowers~\cite{nilsback2008automated} when using our full model with AlexNet~\cite{krizhevsky2012imagenet} as base network.}
        \label{tab:fewshot}
\end{table}
Our scarce-data approach is similar in spirit to the few-shot learning methods~\cite{vinyals2017match,Prototypica2017,munk2017meta}.
For this reason, we propose here a comparison with the state of the art method for few-shot classification, Prototypical networks~\cite{Prototypica2017}.
In the standard few-shot protocol, the task is framed as $N$-way $k$-shot, i.e. provide each time a set of $k$ labeled samples from each of $N$ classes that have not previously been trained upon. 
The goal is then to classify a disjoint batch of unlabeled samples, known  as 'queries', into one of these $N$ classes. 
Therefore, some classes are used to train the few-shot method, while others are only used at test time.
In our case, we do not require such split, as we can train and test the model in all classes simultaneously.
Moreover, their test episodes are composed of only $N$ classes at a time, where $N$ is generally a small number (e.g. below 20).
Contrarily, we follow a more general classification approach and test on all classes simultaneously, which is inherently more challenging as the misclassification probability increases. 

We propose two different scenarios to compare our method to Prototypical networks on the task of Flower~\cite{nilsback2008automated} classification.
The first, \emph{20-way 5-shot}, closely resembles the setting introduced by~\cite{vinyals2017match} and usually employed by few-shot approaches. 
We split the set of classes in train and test, selecting 20 random classes for the testing phase.
Then, we run Prototypical networks for the 20-way 5-shot classification task, following similar settings to those used in the mini-ImageNet experiment of~\cite{Prototypica2017}.
We train until convergence using 100 training episodes and test using 5 episodes, with 5 queries per episode both during training and testing.
The second scenario, \emph{102-way 5-shot}, is more similar to the conventional classification task, in which all classes are used for training and testing.
We maintain the training settings for this case, but remove from the `shot' set those queries used at test time.
Table~\ref{tab:fewshot} presents the results of these experiments.
Our method leads to substantially superior performance in both cases, but the difference is especially remarkable for the 102-way setting.
This demonstrates the limitations of this type of few-shot approaches when scaling to many classes, even when they are trained with the same set of classes used for test. 

\section{Conclusions}
\label{sec:conclusions}
In this paper, we investigated the role of saliency in improving the classification accuracy of a CNN when the available training data is scarce.  For that purpose we have considered adding a saliency branch to an existing CNN architecture, which is used to modulate the standard bottom-up visual features from the original input image. We have shown that the proposed approach leads to an improvement of the recognition accuracy with limited number of training data, when applied to the task of fine-grained object recognition. 

Extensive evaluation has been performed on several datasets and under different settings, demonstrating the usefulness of saliency for fine-grained object recognition, especially for the case of scarce training data. 
In addition, our approach allows to compare saliency methods on the high-level task of fine-grained object recognition. Traditionally, saliency methods are evaluated on their ability to generate saliency maps that indicate the relative relevance of regions for the human visual system. However, it remained unclear if these saliency methods would actually translate into improved high-level vision results for tasks  such as object recognition. Our experiments show that there exists a clear correlation (Pearson product-moment correlation coefficient of 0.95) between the performance of saliency methods on standard saliency benchmarks and the performance gain that is obtained when incorporating them in a object recognition pipeline.
Future work will be devoted to extend the current framework by proposing an end-to-end deep architecture that estimates automatically the saliency map, thus eliminating the need for pre-computing it off-line.

\section*{Acknowledgements}
This work is partially funded by MINECO grant TIN2016-79717-R, Spain. Carola Figueroa is supported by a Ph.D. scholarship from CONICYT, Chile. We acknowledge the CERCA Programme of Generalitat de Catalunya. We also acknowledge the generous GPU support from NVIDIA.




\section*{References}


\bibliography{refs}

\end{document}